\documentclass[10pt,twocolumn,letterpaper]{article}
\usepackage{iccv}              

\definecolor{iccvblue}{rgb}{0.21,0.49,0.74}
\usepackage[pagebackref,breaklinks,colorlinks,allcolors=iccvblue]{hyperref}
\usepackage{dirtytalk}
\usepackage{microtype}

\def\paperID{3} 
\def\confName{Workshop on Responsible Imaging}
\def\confYear{2025}

\title{Sensitivity, Specificity, and Consistency: A Tripartite Evaluation\newline of Privacy Filters for Synthetic Data Generation}

\author{
Adil Koeken, Alexander Ziller, Moritz Knolle, Daniel Rueckert\\
Chair for AI in Healthcare and Medicine, \\Technical University of Munich (TUM) and TUM University Hospital\\
Ismaninger Str. 22, 81675 Munich, Germany\\
{\tt\small adil.koeken@tum.de}
}

\begin{document}
\renewcommand{\paragraph}[1]{\noindent\textbf{#1} \quad}
\maketitle
\begin{abstract}
The generation of privacy-preserving synthetic datasets is a promising avenue for overcoming data scarcity in medical AI research. 
Post-hoc privacy filtering techniques, designed to remove samples containing personally identifiable information, have recently been proposed as a solution. 
However, their effectiveness remains largely unverified.
This work presents a rigorous evaluation of a filtering pipeline applied to chest X-ray synthesis. 
Contrary to claims from the original publications, our results demonstrate that current filters exhibit limited specificity and consistency, achieving high sensitivity only for real images while failing to reliably detect near-duplicates generated from training data. 
These results demonstrate a critical limitation of post-hoc filtering: rather than effectively safeguarding patient privacy, these methods may provide a false sense of security while leaving unacceptable levels of patient information exposed.
We conclude that substantial advances in filter design are needed before these methods can be confidently deployed in sensitive applications.
\end{abstract}

\section{Introduction}
Artificial intelligence (AI) systems rely on large, well-curated datasets.  While such data are indispensable for training accurate models, in sensitive fields such as medicine, they also contain sensitive personal information that must be protected against inadvertent disclosure. To this end, differential privacy (DP)  \citep{dwork2006differential} has become the gold standard for providing provable guarantees that limit the success of privacy attacks \citep{abadi2016deep, song2013stochastic}.  However, DP introduces a well‑known privacy-utility trade‑off \citep{de2022unlocking}, specifically for generative models.
Thus, researchers are exploring alternatives that preserve model performance while still safeguarding patient privacy.

A proposed alternative is the generation of synthetic datasets that retain the statistical properties required for downstream learning but prevent patient re-identification. Recent work has explored two complementary strategies for the generation of synthetic, privacy-preserving medical image datasets: (i) integrating privacy constraints directly into the training objective \citep{pennisi2023privacy}, and (ii) applying post‑hoc filters to remove or modify samples deemed too similar to any real patients' data \citep{reynaud2024echonet}. 
Here, we focus on the latter approach by \citet{reynaud2024echonet}, who propose a latent‑space filtering pipeline for echocardiogram video synthesis. 
Their method trains a contrastive encoder to cluster video frames from the same patient and separate those from different patients; a distance threshold derived from a reference set then determines whether a generated frame is rejected as privacy‑critical.

\begin{figure}[t]
    \centering
    \includegraphics[width=1\linewidth]{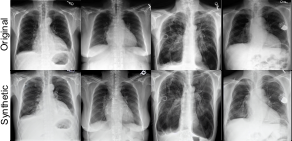}
    \caption{\textbf{Privacy filters fail to detect near-duplicates of training images produced by a diffusion model.} 
    Panels show exemplary failure cases where the privacy filter failed: original training images (top row), synthetic images produced by the diffusion model (bottom row).
    All four images contain distinct visual cues that would allow for patient re-identification.
    }
    \vspace{-5mm}
    \label{fig:leaky_grid_selected}
\end{figure}

In this paper, we rigorously assess the effectiveness of a privacy filter when applied to a state-of-the-art chest X-ray generation model.
Our results reveal that current filters exhibit limited specificity, sensitivity, and consistency, failing to reliably detect near-duplicates (see \cref{fig:leaky_grid_selected}). This poses a significant security risk to patient privacy. 
In this work, we propose a systematic adaptation of the \citet{reynaud2024echonet} latent‑space filter to image‑level data and an additional pixel‑space variant.
Furthermore, we propose an evaluation protocol that measures sensitivity, specificity, and inter‑model consistency on both real held‑out images and diffusion‑generated near‑duplicates.
Following this protocol, we provide empirical evidence that current privacy filter pipelines: 
\begin{itemize}
    \item achieve high sensitivity only for actual images of the same patient,
    \item suffer a substantial drop for synthetic near‑duplicates,
    \item exhibit limited specificity, and display virtually no agreement across random seeds.
\end{itemize}
Together, these findings underscore the need for more reliable techniques to generate privacy-preserving synthetic data.

\section{Background}

\citet{reynaud2024echonet} propose a protocol to generate privacy-preserving synthetic datasets with a focus on video generation (echocardiogram videos). 
Central to their approach is a privacy filtering module, which is based on a neural network image encoder. 
Specifically, this image encoder is trained using contrastive learning to \say{pull-together} (in latent space) video frames from the same patient and \say{push apart} frames from different patients.
In essence, their privacy filter flags generated synthetic images which are \say{too similar} in feature space to any given training image.
Similarity between feature vectors is measured by the Pearson correlation coefficient.

To arrive at a flagging decision, an additional calibration step (originally proposed by \cite{dar2025unconditional}) is required to determine a threshold $\tau$, by which images are flagged/accepted.
The calibration procedure uses a reference dataset which contains all training images and an additional set of validation images.
Concretely, the procedure involves calculating the pair-wise correlation scores for all of the closest training-validation image pairs (closest here refers to max-aggregation of scores over all training images).
The decision threshold $\tau$ is then calculated as the $95$th percentile of these values.
In practice, generated images with a correlation score above the value $\tau$ to any training image are flagged as privacy-critical, and those below are accepted.

\section{Methods}
\subsection{Evaluation}
In our work, we evaluate the protocol of \citet{reynaud2024echonet} for privacy filtering. 

\paragraph{Privacy Filter Desiderata}
We begin by proposing three basic desiderata that a privacy filter should simultaneously satisfy. To provide meaningful privacy protection and utility, a privacy filter should be:
\begin{enumerate}
    \item \textbf{Sensitive:} flag images containing information from patients in the training set (high true-positive rate).
    \item \textbf{Specific:} avoid flagging images from patients not part of the training set (low false‑positive rate).
    \item \textbf{Consistent:} produce concordant decisions on identical inputs, regardless of random seed.
\end{enumerate}

\paragraph{Modifications}
We focus solely on privacy filtering and are not interested in video generation in this work. 
For this reason, we do not consider videos but image datasets. 
The protocol by \citet{reynaud2024echonet} remains largely unmodified, except that we cannot perform a contrastive training of frames from the same video. 
Instead, we use images of the same patients as positive pairs and images of different patients as negative pairs. 

\citet{reynaud2024echonet} apply their privacy filter directly in the latent space of a Variational Autoencoder (VAE).
We extend this approach and additionally evaluate how effectively a privacy filter can operate in pixel space, directly on the high-resolution synthetic images.

\paragraph{Sensitivity}
\citet{reynaud2024echonet} evaluate the quality of a privacy filter by assessing the accuracy with which frames from training and validation videos can be re-assigned. 
This aligns with the aim of detecting samples with \say{revealing} patient attributes. 
We adapt these experiments to our settings by leveraging patients with multiple images available. 
Here, we put half of their images as training images in the reference set. 
Then we present the other half of the images as synthetic data to be filtered and evaluate the sensitivity of the filters in flagging these images.
The same-patient strategies isolate the filter's response to genuine matches (images from the same patient), providing a measure of the filter's sensitivity when true positives are guaranteed.

In an ideal privacy filter, the flagging decision should depend solely on whether similar patient characteristics exist in the training pool.
To evaluate this, we compared three flagging strategies that vary the comparison pool (reference training images):

\begin{itemize}
    \item \textbf{Overall}: holdout presented as synthetic images were compared against the combined training pool (reference training + holdout training)
    \item \textbf{Same-patient (maximum)}: For each test image, we computed correlations only with training images from the same patient.
    \item \textbf{Same-patient (mean)}: Similar to above, but flagging was based on the mean correlation with same-patient training images exceeding $\tau$ instead of the max correlation setting.
    This makes the flagging threshold independent of the number of reference images, which disproportionately affects patients with multiple images.
\end{itemize}
The same-patient strategies isolate the filter's response to genuine matches (images from the same patient), providing a measure of the filter's sensitivity when true positives are guaranteed. 
However, it is not clear if synthetically generated images follow the same patterns. 
To investigate this, we additionally generate \say{synthetic ground truth samples}. 
\citet{carlini2023extracting} showed that text-conditioned diffusion models have a tendency to generate near duplicates of training images if they are conditioned on the same text input as was used during model training. 
We exploit this property to generate synthetic images, which are very similar to a training sample and evaluate their detection of the privacy filters. 

\paragraph{Specificity}
While sensitivity is the most important criterion for effective privacy filtering, specificity is crucial for making a privacy filter useful in practice. High specificity ensures that filters \say{understand} intra-patient features and, by that, prevent unnecessary rejection of synthetic images. This also keeps computation cost manageable while maintaining dataset diversity.
We evaluate this by presenting the filter with real images of patients not part of the training set and checking if they are flagged as privacy-critical. This ensures that holdout images have no genuine patient matches in the training pool. Since these images cannot contain information from any training patient, any flagged image represents a false positive. This setup directly measures the filter's ability to distinguish between genuine privacy risks and spurious inter-patient similarities.

\paragraph{Consistency}
As a last criterion, we postulate that well-functioning privacy filters are consistent in the same setting. 
Specifically, their decision whether a synthetic image is flagged as privacy critical should be independent of the random seed used in training the filter. 
To evaluate this, we train $N$ privacy filters with different random seeds and evaluate their agreement in flagging generated synthetic images. 

\section{Results}

 \begin{figure*}[tp]
     \centering
    \includegraphics[width=0.9\linewidth]{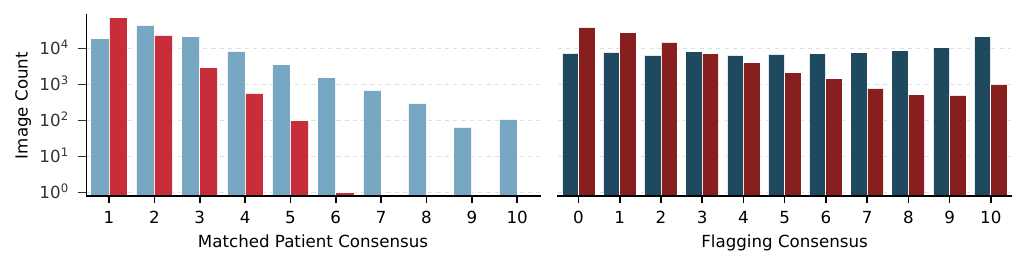}
     \caption{\textbf{Privacy filters derived from different random seeds are inconsistent in their decisions.}
     We compare the outputs of $10$ distinct privacy filters derived from different random seeds when applied to the same set of $100\,000$ synthetic images.
     Sub-panels show the level of consensus (i.e., number of images for which $N$ out of $10$ filters agree) for patient attribution (left) and binary filtering decisions (right).
     Left blue bars show results for filters in pixel space, right red bars show results in latent space. 
     }
     \label{fig:consensus}
\end{figure*}

\paragraph{Setup and Dataset}
We evaluate privacy filters with RoentGen \citep{chambon2022roentgen}, a state-of-the-art chest X-ray generation model trained on MIMIC-CXR \citep{johnson2019mimic}. 
The availability of other datasets within the same domain, in particular CheXpert \citep{irvin2019chexpert}, allow us to evaluate privacy filters also on unseen data. 

The privacy filters were trained following the exact protocol of \citet{reynaud2024echonet} (including hyperparameters).
Similar to their work, we achieve high classification metrics on held-out data: AUC=$0.9993$, Recall=$0.9907$, Precision=$0.9873$.

\paragraph{Sensitivity}
\begin{table}[t]
\centering
\begin{tabular}{lcc}
\toprule
\textbf{Compared Set} & \textbf{Pixel (\%)} & \textbf{Latent (\%)} \\
\midrule
Overall (combined pool) & $88.5$ & $75.8$ \\
Same-patient (maximum) & $59.8$ & $8.7$ \\
Same-patient (mean) & $24.2$ & $0.9$ \\
\bottomrule
\end{tabular}
\caption{Flagging rates for same-patient positive pair analysis across different filtering pool strategies}
\label{tab:sensitivity_analysis}
\end{table}
Table~\ref{tab:sensitivity_analysis} shows flagging rates when holdout test images from patients with known training matches are evaluated under different comparison strategies. When test images were compared against the combined training pool (Overall), the filters flagged $88.5\%$ of images in pixel space and $75.8\%$ in latent space, suggesting high sensitivity at first glance. However, this apparent sensitivity is misleading.

When we isolate comparisons to same-patient training images -- where true matches are guaranteed -- flagging rates drop precipitously. Using maximum correlation within same-patient matches, only $59.8\%$ (pixel) and $8.7\%$ (latent) of images exceed the threshold $\tau$. With mean correlation, these rates further decrease to $24.2\%$ and $0.9\%$, respectively. This dramatic reduction reveals that the filters' high overall flagging rate stems primarily from spurious matches to different patients rather than accurate identification of same-patient similarities.

To further assess sensitivity on synthetic data, we generated ground truth samples using the approach of \citet{carlini2023extracting}, prompting the diffusion model with exact training prompts to intentionally produce near-duplicates. 
In this setting only $8\,656$ ($8.7\%$) successfully matched their source prompt's patient ID -- $7\,881$ flagged as privacy-critical and $775$ unflagged. Among the correctly matched but unflagged images, manual inspection revealed numerous visually similar pairs (Figure~\ref{fig:leaky_grid_selected}), demonstrating that the filters fail to detect even intentionally generated near-duplicates -- precisely the scenario they were designed to prevent.

\paragraph{Specificity}
When evaluating holdout images from patients entirely absent from the training set, the privacy filters demonstrated poor specificity.  The filters incorrectly flagged $84.3\%$ of holdout images in pixel space and $76.8\%$ in latent space as privacy-critical. These excessive false positive rates indicate that the filters cannot reliably distinguish between actual privacy violations and benign inter-patient similarities. In practical deployment, such indiscriminate rejection would eliminate the vast majority of safe synthetic samples, rendering the generation of a large synthetic dataset prohibitively computationally expensive.

\paragraph{Consistency}
We analyzed the level of agreement among $N=10$ independently trained filters evaluated on an identical set of $100\,000$ synthetic images. Figure~\ref{fig:consensus} reveals substantial inconsistency: latent space filters achieve unanimous consensus on only $40.4\%$ of images ($39\,341$ safe, $1\,022$ flagged), while pixel space filters reach agreement on $29.0\%$ ($7\,534$ safe, $21\,453$ flagged). The majority of images receive contradictory assessments across different model initializations.

Among $100\,000$ generated images, unanimous filtering decisions were rare: in latent space, only $193$ images were unanimously flagged as privacy-critical across all 10 filters, with $12\,924$ unanimously deemed safe. Pixel space showed higher but still inadequate consensus, with $30\,730$ unanimously flagged and $1\,332$ unanimously safe.

Patient attribution consistency proves even more problematic (Figure~\ref{fig:consensus}). For $72\,466$ images in latent space and $19\,047$ in pixel space, no two filters agreed on the attributed patient -- each filter linked the image to a different training subject. Complete consensus across all filters occurred for only $110$ images in pixel space and never in latent space, where maximum agreement was limited to six filters.

Notably, we found that, even when filters unanimously agree on patient attribution, they frequently disagree on whether to flag the image as privacy-critical. We thus hypothesize that filtering decisions are dominated by random initialization effects rather than deterministic privacy characteristics.

\section{Discussion}

This work presents a critical assessment of AI privacy filters as a means to generate privacy-conforming synthetic datasets. Our analysis centers around three key principles: sensitivity -- the ability to correctly identify images containing patient data from the training set; specificity -- ensuring that images from unseen patients are not incorrectly flagged; and consistency -- achieving reliable, reproducible results across different random seeds. We find that while current privacy filters demonstrate reasonably high flagging rates on real images of patients contained in the training set, this performance is accompanied by several important limitations.
Specifically, we observed that synthetic near-duplicates of training images are not consistently detected by these filters. This is particularly concerning given recent demonstrations \citep{carlini2023extracting} of how text-conditioned diffusion models can be prompted to reproduce training examples and how much unconditional diffusion models memorize training data \citep{dar2025unconditional}. Furthermore, we found that even real images originating from patients not present in the training dataset are flagged at a surprisingly high rate. This suggests that current filters struggle to differentiate between true privacy risks and benign similarities between patients.

The lack of consistency across independently trained filters further complicates matters. We observed substantial disagreement both in assigning synthetic images to potential patient identities within the training set and in determining whether a correctly assigned image should be considered privacy-critical. This indicates that the detection performance of these filters may rely heavily on excessive filtering -- flagging many images as potentially sensitive even when they pose no actual risk -- rather than learning robust, discriminative intra-patient features. A well-grounded privacy filter should ideally identify and protect against specific patient characteristics, not simply err on the side of caution by rejecting a large proportion of generated data.


Our study does not come without limitations. Firstly, our evaluation was conducted using a single imaging dataset. A more comprehensive assessment would involve testing multiple filters across a wider variety of medical image modalities and datasets. Secondly, while we leveraged prompt engineering techniques based on the work of \citet{carlini2023extracting} to generate synthetic \say{ground truth} samples, it's possible that even these near-duplicates do not fully capture all potential privacy vulnerabilities. Finally, our study largely fixed the size of the reference dataset; recent theoretical research has shown a strong influence of data pool size on filter performance \citep{rocher2025scaling}, and future investigations should explore this relationship in more detail.

In summary, our results serve as an indication that current privacy filtering methods fall short of providing the protection that was previously believed.
We hope that this work can stimulate further research to enhance the effectiveness of privacy filters, which we hope could ultimately lead to the holy grail of AI: large representative training datasets stripped of sensitive private information.

\section*{Acknowledgements}
This work was supported by the German Federal Ministry of Research, Technology, and Space (BMFTR) under grant number 01ZZ2316C (PrivateAIM) and and under the DAAD programme Konrad Zuse Schools of Excellence in Artificial Intelligence.


{
    \small
    \bibliographystyle{ieeenat_fullname}
    \bibliography{main}
}

\end{document}